# Generative and Discriminative Deep Belief Network Classifiers: Comparisons Under an Approximate Computing Framework

Siqiao Ruan[1], Ian Colbert[2], Ken Kreutz-Delgado[2], and Srinjoy Das[3][**]

*Abstract*—The use of Deep Learning hardware algorithms for embedded applications is characterized by challenges such as constraints on device power consumption, availability of labeled data, and limited internet bandwidth for frequent training on cloud servers. To enable low power implementations, we consider efficient bitwidth reduction and pruning for the class of Deep Learning algorithms known as Discriminative Deep Belief Networks (DDBNs) for embedded-device classification tasks. We train DDBNs with both generative and discriminative objectives under an approximate computing framework and analyze their power-at-performance for supervised and semi-supervised applications. We also investigate the out-of-distribution performance of DDBNs when the inference data has the same class structure yet is statistically different from the training data owing to dynamic real-time operating environments. Based on our analysis, we provide novel insights and recommendations for choice of training objectives, bitwidth values, and accuracy sensitivity with respect to the amount of labeled data for implementing DDBN inference with minimum power consumption on embedded hardware platforms subject to accuracy tolerances.

## I. Introduction and Background

Discriminative Deep Belief Networks (DDBNs) are stochastic deep neural networks that have been used in a wide range of applications [1]-[3]. A DDBN is a stochastic neural network that extracts a deep hierarchical representation from data. It can be trained in a greedy fashion by sequentially learning Restricted Boltzmann Machines (RBMs) [5] using the architecture shown in Fig. 1(a). Afterwards, classification with the trained DDBN can be performed by estimating the values of neurons in each layer using probabilistic inference on the architecture shown in Fig. 1(b). The stochastic nature of a DDBN allows it to be trained using both generative and discriminative training objectives, supporting supervised and semi-supervised tasks. Given inputs $x$ and labels $c$ a generative learning objective can be used to train the weights of a DDBN by performing maximum likelihood estimation (MLE) over the joint distribution $p(x,c)$. It is also possible to train a DDBN with a discriminative objective by performing MLE over the conditional probability distribution $p(c|x)$. Generative and discriminative learning objectives for classifiers have been discussed previously in [9] and analyzed specifically for Discriminative RBMs in [4].

In recent years, there has been a proliferation of Deep Learning applications for performing classification tasks on embedded computing processors on the edge [10]-[13]. Several approximate computing frameworks [14][15] for embedded deployment have been proposed for feedforward neural networks and convolutional neural networks, but little work has been done on stochastic deep networks like DDBN.

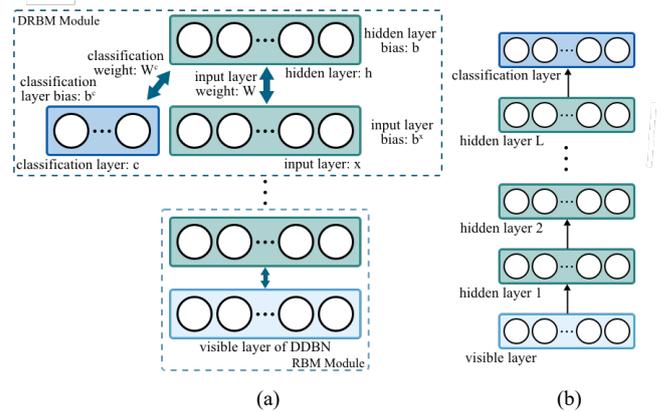

Fig. 1. (a) DDBN architecture using stacked RBMs (b) Inference process of DDBN with L hidden layers.

For efficiently realizing classification with DDBNs on embedded platforms, algorithm designers must take several critical considerations into account. These include amount of maximum possible bitwidth reduction for target accuracy, robustness of classification performance, and limited availability of labeled data. To address these factors, we expand on an approximate computing methodology (AX-DBN) first proposed in [6] and perform an analysis of both supervised and semi-supervised applications for approximated DDBNs using both generative and discriminative training objectives. We empirically demonstrate the impact of these training methodologies and the amount of labeled data on obtained bitwidth distributions subject to user specified tolerances with respect to full precision classification accuracy. The AX-DBN methodology involves training and approximation on cloud servers for deployment on a local device. In such cases, the user may not always have access to stable internet connectivity for frequent training and the operating environment where inference is performed may be dynamically changing. Taking such constraints into account, we also analyze the classification performance of DDBNs with out-of-distribution samples where the inference data has the same class structure but different statistics with respect to the training data.[1] Taking all these factors into consideration and going beyond the standard results for the full precision networks discussed in [4][5], we provide a comprehensive analysis with key insights and recommendations for efficient inference with approximated DDBNs on embedded platforms with minimal power consumption under user specified accuracy tolerances.

## II. Approximate Computing Framework for DDBN

To meet the constraints of low power consumption during inference, we enable sparse implementations of DDBNs using AX-DBN which is an approximate computing methodology proposed by [6]. Given a maximum tolerable loss w.r.t full precision classification accuracy and a set of bitwidths, the AX-DBN framework is used to obtain a bit-reduced model using a set of greedy iterations involving neuron pruning and network retraining. It performs a heterogeneous bitwidth approximation where each of the weights and biases of the final model can individually take any value from the set of user-specified bitwidth values. In this methodology (Fig. 2), the first step involves performing a uniform bitwidth reduction across all neurons of the full precision model until the user-specified minimum accuracy tolerance is violated. Then, neurons are ranked using the cross-entropy loss function to determine their criticality on classification performance. The least critical neurons are pruned to the next available lower bitwidth followed by which the network is retrained to maximize its performance with the reduced bitwidths. These criticality-retraining iterations are performed on the cloud server upon request from the local embedded device until the bitwidths of the model cannot be reduced any further. Afterwards, the approximated model is deployed on the local device for performing low power inference. Previously AX-DBN was used to obtain a sparse DDBN model by using the generative training objective. Here, we incorporate the discriminative objective and analyze AX-DBN for semi-supervised learning which dominates real-world classification applications. Practical constraints on internet bandwidth dictate that the local device may not always be able to request retraining (Fig. 2) to adjust to changes in environment. Therefore, we also analyze the robustness of classification using out-of-distribution samples when the network is trained using both objectives in order to provide algorithmic insights for implementing DDBNs on low-power devices.

[1]Out-of-distribution data can arise in a gradually changing operating environment during real-time applications.

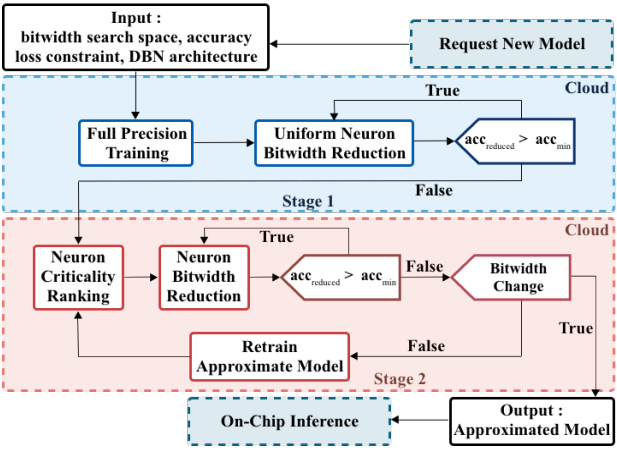

Fig. 2. AX-DBN Methodology Diagram [6]

## III. DISCRIMINATIVE DBNs AND LEARNING OBJECTIVES

The building blocks of DDBNs are RBMs (Restricted Boltzmann Machines). An RBM is a stochastic neural network of two interconnected layers consisting of visible neurons $v$ and hidden neurons $h$. For classification tasks, a Discriminative RBM (DRBM) can be formed by dividing the visible layer of an RBM into two parts: inputs $x$ and its associated "one-hot" class $c$ [4]. The DRBM captures a probabilistic generative model of the input data based on the Boltzmann distribution as below:

$$P(x,c,h) = \frac{e^{-E(x,c,h)}}{Z} \text{ where } Z = \sum_{x,c,h} e^{-E(x,c,h)} \quad (1)$$

where $E(x,c,h) = -x^T b^x - h^T b - c^T b^c - x^T W h - c^T W^c h$

Here $P$ denotes the Boltzmann probability distribution, and $E$ is an energy function of $x$, $c$ and $h$. The terms $b^x$, $b$ and $b^c$ are biases of the visible, hidden, and classification layers respectively. $W$ and $W_c$ are weight matrices between layers as shown in Fig. 1(a). A DDBN is formed by stacking RBMs and concatenating a classification layer to the final hidden layer as shown in Fig. 1. It was shown in [5] that DBNs can be trained by sequentially learning RBMs. For inference, the activation probability of the neurons on each hidden layer can be determined successively based on the given data. Once the activation value of the last hidden layer $L$ is determined then the Free Energy corresponding to each class label $c$ can be calculated from:

$$F(x, c_i = 1) = -b_i^c - \sum_{j=1}^{H} \log(1 + e^{b_j + W_{ji}^c + W_j^T x}) \quad (2)$$

The classification result is given by the class c that corresponds to the minimum Free Energy.

It was shown in [4] that for both supervised and semi-supervised tasks, a DRBM can be trained using both generative and discriminative learning objectives. Given a labeled training dataset $D_{labeled} = \{(x^1, c^1), ..., (x^N, c^N)\}$, the generative learning objective and discriminative learning objective for DRBM is defined as follows:

$$L_{gen}(D_{labeled}) = -\frac{1}{N} \sum_{n=1}^{N} \log(P(x = x^n, c = c^n)) \quad (3)$$

$$L_{disc}(D_{labeled}) = -\frac{1}{N} \sum_{n=1}^{N} \log(P(c = c^n \mid x = x^n)) \quad (4)$$

For learning, we calculate the derivative of $log(P(x=x^n, c=c^n))$ and $log(P(c=c^n|x=x^n))$ with respect to the parameters $\theta \in \{W, W^c, b^x, b^c, b\}$ as in the description in [4]. We extend this DRBM learning framework to DDBNs, which consists of stacked RBMs. DDBNs trained in this manner with generative and discriminative objectives are denoted as GT-DBN and DT-DBN respectively.

## IV. SEMI-SUPERVISED LEARNING WITH DDBN

For many real-world applications, the available data is a mix of unlabeled ($D_{unlabeled} = \{u^1, ..., u^U\}$) and labeled ($D_{labeled}$) components. In this case, $D_{unlabeled}$ can be used by the DDBN to learn the representation of the underlying data [7][17] which can be used in conjunction with $D_{labeled}$ for efficient classification. Using the labeled data, we can train the DDBN using either the generative ($L_{gen}$) or discriminative training ($L_{disc}$) objectives which were defined previously in Section III. The overall semi-supervised learning objective using both labeled and unlabeled components of the data is given as [4]:

$$L_{ssl}(D_{labeled}, D_{unlabeled}) = L_{\{gen,disc\}}(D_{labeled}) + \beta L_{unsup}(D_{unlabeled}) \quad (5)$$

Here $\beta$ is a hyperparameter and the unsupervised loss for unlabeled data is defined as below:

$$L_{unsup}(D_{unlabeled}) = -\frac{1}{U} \sum_{n=1}^{N} \log(P(x = u^n)) \quad (6)$$

## V. EXPERIMENTAL SETUP AND COMPARISON METRICS

We perform a set of empirical analysis in order to study the effect of training objectives, amount of unlabeled samples, and out-of-distribution data on classification performance using the setup as described hereafter. Our dataset for all experiments consists of 60000 training images and 10000 test images from the MNIST dataset. For supervised classification, all training images are assumed to be labeled whereas, in the semi-supervised case, we consider a mix of m labeled and n unlabeled data. In this paper, [l, u] combinations used are [10000, 50000], [30000, 30000], and [50000, 10000]. Our simulations are performed using two architectures: DBN-200-100 and DBN-300-200-100 where the numbers indicate the number of hidden units in each RBM submodule which are the building blocks of the DDBNs. For cloud-based training and on-chip inference in the AX-DBN framework we consider tolerances of 1% and 5% with respect to full precision accuracy and a search space of 0 (neuron fully pruned), 4-bits Q(2.2), 8-bits Q(2.6), 12-bits Q(6.6), 16-bits Q(8.8), 32-bits Q(8.24) and 64 bits Q(8.56) bits where Q(m.n) denotes m integer bits and n fractional bits.[3] For comparison of different algorithms (GT-DBN vs. DT-DBN, supervised vs. semi-supervised), we perform 200 Monte Carlo (MC) training iterations, each run randomly initializing the weights and biases of the network. The mean bitwidths, mean full-precision (FP) model accuracies, and mean bitwidth-reduced (BR) model accuracies are obtained over all MC iterations for the test data. For measuring out-of-distribution performance, we add salt and pepper noise to the original test data with noise factors equal to 0.1, 0.2 and 0.3 (Fig. 3) and obtain mean classification accuracies over all runs. Unlike test data which has the same statistics as the training data, such out-of-distribution samples have different statistics however retain the same class structure for purposes of discrimination w.r.t both training and test data.

---
[2]The value β = 0.1 is used in our experiments. To choose β, we perform grid-search over β between {0.001, 0.1} to obtain the β that corresponds to the best classification accuracy using test data.

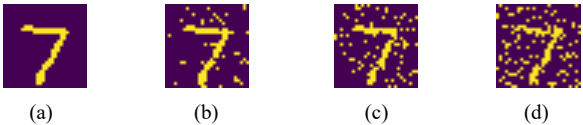

Fig. 3. (a) MNIST test data. (b)-(d) MNIST test data with salt & pepper noise 0.1, 0.2, and 0.3.

## VI. COMPARISON OF LEARNING OBJECTIVES

Our first comparison between GT-DBN and DT-DBN is done for supervised classification. The results are summarized in Table I. The full precision accuracy of DT-DBN is higher than GT-DBN which is consistent with [4]. However, when accuracy tolerances are specified under the AX-DBN framework, the network is pruned less using the discriminative versus generative training objective as seen from the mean bitwidths. To investigate further, we restricted the search space of AX-DBN to include only {0,4,8} bits with an accuracy tolerance of 5%. The results over 200 MC training iterations for both DT-DBN architectures are shown in Fig. 5. The vertical red line in all plots denote the minimum accuracy of the pruned network w.r.t the highest full-precision accuracy for the given accuracy tolerance. In this case, for each MC iteration, most of the neurons are pruned to 8 bits and the specified accuracy tolerance is never met. In contrast, when trained with GT-DBN using the same limited bitwidth search space, the model achieves the specified accuracy tolerance as shown in Fig. 4. We conclude that discriminative training achieves better full precision accuracy with respect to generative training, but it results in "excess network capacity" i.e. higher overall bitwidth given a lower limit on the amount of pruning specified in terms of a maximum tolerance allowed from the achieved full precision accuracy.

| Architecture (1%) | Generative | | | Discriminative | | |
|---|---|---|---|---|---|---|
| | FP Acc | BR Acc | Avg. Bitwidth | FP Acc | BR Acc | Avg. Bitwidth |
| DBN-200-100 | 92.75 | 91.96 | 8.06 | 94.51 | 94.27 | 11.94 |
| DBN-300-200-100 | 93.32 | 92.42 | 9.99 | 94.4 | 94.13 | 11.65 |
| Architecture (5%) | Generative | | | Discriminative | | |
| | FP Acc | BR Acc | Avg. Bitwidth | FP Acc | BR Acc | Avg. Bitwidth |
| DBN-200-100 | 92.71 | 88.49 | 6.82 | 94.51 | 92.60 | 10.77 |
| DBN-300-200-100 | 93.33 | 89.38 | 7.70 | 94.37 | 92.61 | 10.56 |

TABLE I. SUPERVISED GT-DBN AND DT-DBN ACCURACY AND AVG. BITWDITH WITH 1% (TOP) AND 5% (BOTTOM) ACCURACY TOLERANCE.

---

[3]The choice of m and n gives the best accuracy when the model is approximated uniformly across all neurons.

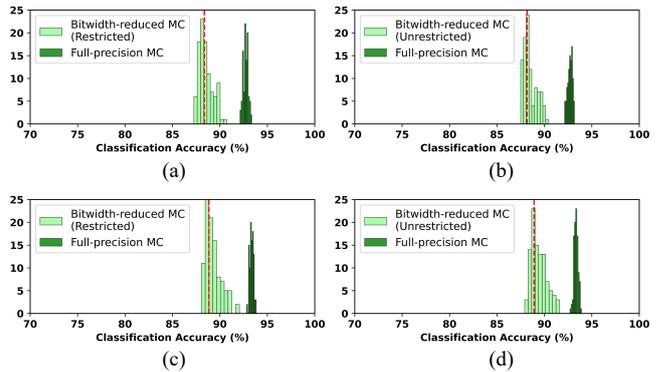

Fig. 4. (a)-(b) Classification Acc distributions of restricted and unrestricted bitwidth search space for GT-DBN-200-100 with 5% accuracy tolerance. (c)-(d) Classification Acc distributions of restricted and unrestricted bitwidth search space for GT-DBN-300-200-100 with 5% accuracy tolerance.

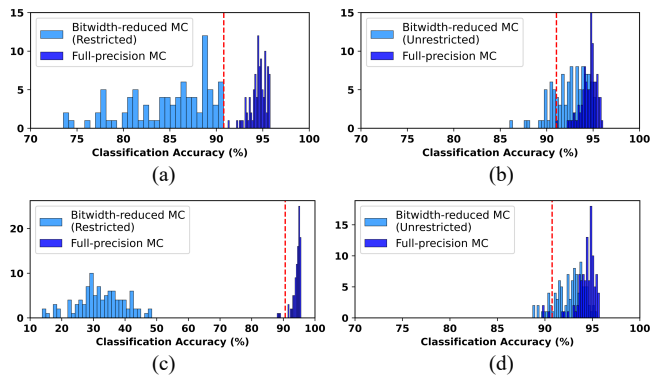

Fig. 5. (a)-(b) Classification Acc distributions of restricted and unrestricted bitwidth search space for DT-DBN-200-100 with 5% accuracy tolerance. (c)-(d) Classification Acc distributions of restricted and unrestricted bitwidth search space for DT-DBN-300-200-100 with 5% accuracy tolerance.

Results for semi-supervised classification are summarized in Table II. In this case, we see that the gap between mean bitwidths for DT-DBN and GT-DBN is reduced as compared to the purely supervised case. This effect is more pronounced for the shallower network. For deeper network, DT-DBN shows some excess capacity after pruning as compared to GT-DBN, however the gap is much narrower as compared to supervised classification. This behavior can be attributed to the unsupervised component of the training objective (6) which enables learning representations from the unlabeled data and regularizes the excess capacity seen with purely labeled data using the discriminative learning objective. As a result, the discriminative training objective in the semi-supervised case achieves a more continuous trend for reduction in mean bitwidth for a given tolerance with respect to full precision accuracy.

## VII. COMPARISON OF OUT-OF-DISTRIBUTION PERFORMANCE

We compare out-of-distribution classification performance using GT-DBN and DT-DBN for both supervised and semi-supervised classification. Results are shown in Fig. 6. Even though DT-DBN has higher full precision accuracy versus GT-DBN which is consistent with [4], it has lower out-of-distribution accuracy on the pruned network. This effect is particularly strong when a higher noise factor is used which results in significant accuracy degradation for DT-DBN. However, when we consider semi-supervised classification the gap in performance for out-of-distribution samples between GT-DBN and DT-DBN is considerably reduced as shown in Fig.7. We attribute this difference to superior generalization properties when unlabeled samples are used for training using both objectives.

| Architecture (1%) | Generative | | | Discriminative | | |
|---|---|---|---|---|---|---|
| | FP Acc | BR Acc | Avg. Bitwidth | FP Acc | BR Acc | Avg. Bitwidth |
| DBN-200-100 [10k, 50k] | 91.01 | 90.17 | 8.28 | 92.02 | 91.19 | 8.36 |
| DBN-200-100 [30k, 30k] | 91.88 | 91.04 | 8.40 | 93.48 | 92.67 | 8.43 |
| DBN-200-100 [50k, 10k] | 92.04 | 91.09 | 8.43 | 93.94 | 93.17 | 8.47 |
| DBN-300-200-100 [10k, 50k] | 91.44 | 90.65 | 9.58 | 92.47 | 91.90 | 10.18 |
| DBN-300-200-100 [30k, 30k] | 92.18 | 91.33 | 9.99 | 93.97 | 93.16 | 10.20 |
| DBN-300-200-100 [50k, 10k] | 92.30 | 91.46 | 10.19 | 94.30 | 93.46 | 10.50 |

| Architecture (5%) | Generative | | | Discriminative | | |
|---|---|---|---|---|---|---|
| | FP Acc | BR Acc | Avg. Bitwidth | FP Acc | BR Acc | Avg. Bitwidth |
| DBN-200-100 [10k, 50k] | 91.04 | 86.90 | 6.36 | 92.05 | 87.98 | 6.36 |
| DBN-200-100 [30k, 30k] | 91.90 | 87.53 | 6.57 | 93.51 | 89.37 | 6.51 |
| DBN-200-100 [50k, 10k] | 92.04 | 87.83 | 6.63 | 93.91 | 91.19 | 6.74 |
| DBN-300-200-100 [10k, 50k] | 91.42 | 87.25 | 7.42 | 92.47 | 87.95 | 8.20 |
| DBN-300-200-100 [30k, 30k] | 92.19 | 87.93 | 7.59 | 93.98 | 89.40 | 8.42 |
| DBN-300-200-100 [50k, 10k] | 92.29 | 88.30 | 7.64 | 94.22 | 90.16 | 8.77 |

TABLE II. SEMI-SUPERVISED DDBN ACCURACY AND AVERAGE BITWIDTH WITH 1% (TOP) AND 5% (BOTTOM) ACCURACY TOLERANCE.

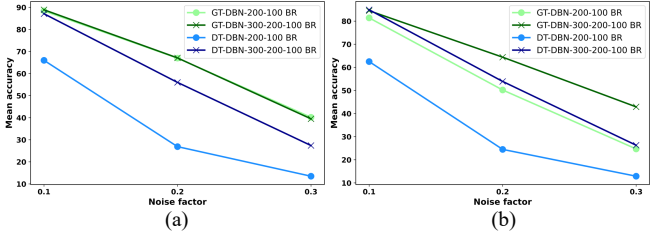

Fig. 6. Supervised DDBN out-of-distribution mean classification accuracy for both full-precision (FP) and bitwidth-reduced (BR) model with (a) 1% accuracy tolerance and (b) 5% accuracy tolerance.

## VIII. EFFECT OF ADDING UNLABELED SAMPLES

Comparing semi-supervised vs purely supervised classification when the number of labeled samples is fixed, we perform a set of experiments using supervised training with 10000 labeled data and semi-supervised learning with 10000 labeled data and 50000 unlabeled data. Results are shown in

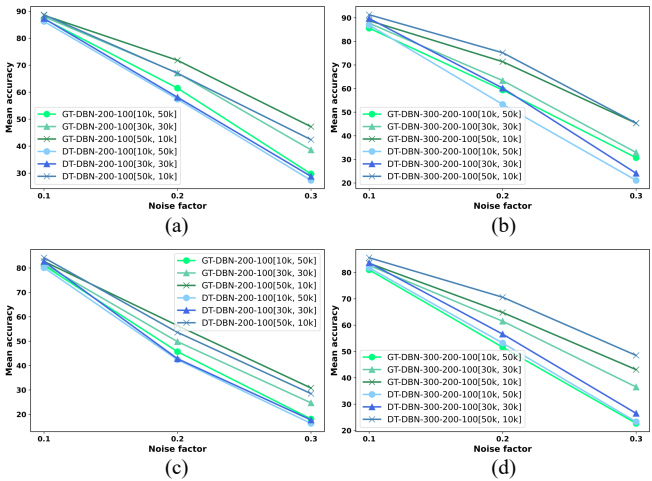

Fig. 7. (a)-(b) Out-of-distribution accuracy of semi-supervised DBN-200-100 and DBN-300-200-100 with 1% accuracy tolerance. (c)-(d) Out-of-distribution accuracy of semi-supervised DBN-200-100 and DBN-300-200-100 with 5% accuracy tolerance.

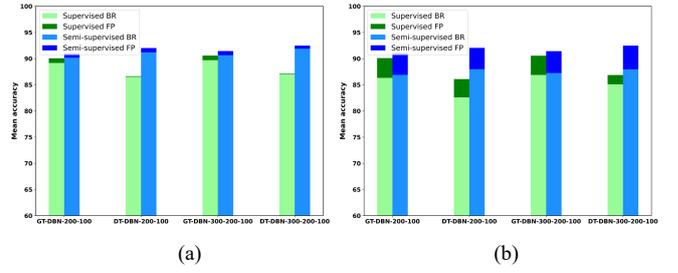

Fig. 8. Supervised [10k] and semi-supervised [10k, 50k] DDBN accuracy with (a) 1% accuracy tolerance and (b) 5% accuracy tolerance. Smaller gap between full-precision (FP) model and bitwidth-reduced (BR) model on a given bar indicates excess network capacity after pruning with AX-DBN.

| Architecture (1%) | Generative Out-of-distribution Accuracy | | | | | | | Discriminative Out-of-distribution Accuracy | | | | | | |
|---|---|---|---|---|---|---|---|---|---|---|---|---|---|---|
| | 0.1 Noise | | 0.2 Noise | | 0.3 Noise | | Avg. bitwidth | 0.1 Noise | | 0.2 Noise | | 0.3 Noise | | Avg. bitwidth |
| | FP | BR | FP | BR | FP | BR | | FP | BR | FP | BR | FP | BR | |
| DBN-200-100 | 87.4 | 85.8 | 68.7 | 60.1 | 41.3 | 27.8 | 8.51 | 59.4 | 54.5 | 26.5 | 22.3 | 13.6 | 12.2 | 11.54 |
| DBN-200-100 (SSL) | 88.8 | 87.1 | 68.9 | 61.5 | 40.7 | 29.7 | 8.43 | 88.8 | 86.2 | 67.1 | 57.4 | 38.4 | 27.3 | 8.43 |
| DBN-300-200-100 | 87.8 | 82.8 | 68.3 | 58.6 | 45.0 | 30.5 | 9.53 | 71.4 | 68.9 | 42.2 | 37.7 | 23.3 | 20.1 | 12.24 |
| DBN-300-200-100 (SSL) | 89.6 | 85.6 | 71.1 | 59.4 | 42.7 | 30.7 | 9.58 | 90.9 | 86.9 | 73.0 | 53.3 | 38.2 | 21.1 | 10.18 |

| Architecture (5%) | Generative Out-of-distribution Accuracy | | | | | | | Discriminative Out-of-distribution Accuracy | | | | | | |
|---|---|---|---|---|---|---|---|---|---|---|---|---|---|---|
| | 0.1 Noise | | 0.2 Noise | | 0.3 Noise | | Avg. bitwidth | 0.1 Noise | | 0.2 Noise | | 0.3 Noise | | Avg. bitwidth |
| | FP | BR | FP | BR | FP | BR | | FP | BR | FP | BR | FP | BR | |
| DBN-200-100 | 87.4 | 81.0 | 68.6 | 45.7 | 41.2 | 18.1 | 6.52 | 59.6 | 49.9 | 27.0 | 20.9 | 14.0 | 12.1 | 9.89 |
| DBN-200-100 (SSL) | 88.8 | 81.5 | 69.0 | 47.3 | 41.2 | 19.1 | 6.36 | 88.8 | 80.1 | 67.1 | 42.4 | 38.6 | 16.3 | 6.36 |
| DBN-300-200-100 | 87.8 | 79.9 | 68.2 | 48.4 | 44.8 | 20.3 | 7.29 | 70.8 | 65.1 | 41.4 | 32.5 | 22.7 | 17.0 | 11.03 |
| DBN-300-200-100 (SSL) | 89.6 | 81.0 | 71.2 | 51.6 | 42.3 | 22.7 | 7.42 | 90.9 | 81.8 | 72.9 | 53.2 | 38.3 | 23.3 | 8.20 |

TABLE III. SUPERVISED [10K] AND SEMI-SUPERVISED [10K, 50K] DDBN OUT-OF-DISTRIBUTION ACCURACY AND AVERAGE BITWIDTH WITH 1% (TOP) AND 5% (BOTTOM) ACCURACY TOLERANCE

Fig. 8 where it is shown that that unlabeled data enhances the ability of our models to utilize the specified accuracy tolerance and thereby avoid excess network capacity. The comparison in Table III shows that in contrast with supervised training, the gap in mean bitwidth and out-of-distribution performance between GT-DBN and DT-DBN is reduced with the addition unlabeled sample. These results are consistent with our findings in Section VI and shows that unlabeled data add more representation capability to both GT-DBN and DT-DBN which results in better performance vs using only labeled data using out-of-distribution samples.

## IX. CONCLUSIONS

Building on preliminary work given in [6], we analyze the important problem of obtaining finite-precision approximate models of a DDBN, a stochastic neural network, given user-specified accuracy fall-off tolerances. We provide an empirical demonstration of the tradeoff between generative and discriminative learning objectives for DDBNs under an approximate computing framework. Our results show the following for approximated DDBNs:

a) For purely supervised classification DT-DBN shows higher mean bitwidths and poor out-of-distribution performance versus GT-DBN.

b) Both mean bitwidths and out-of-distribution accuracy metrics show less sensitivity to the training methodology for semi-supervised classification vs the purely supervised case.

c) For semi-supervised classification both mean bitwidths and out-of-distribution accuracy metrics improve with increasing amount of unlabeled data versus using purely supervised classification with only labeled samples.

Our insights in this paper are meant to provide guidance for algorithm designers on critical metrics such as mean bitwidth and out-of-distribution classification accuracy in order to achieve efficient low-power inference on such platforms. Future work will focus on extending the investigation in this paper to models such as LSTM [16] for other inference tasks on low-power embedded platforms.